%% file: main.tex
\documentclass[conference]{IEEEtran}
\IEEEoverridecommandlockouts
\usepackage{cite}
\usepackage{amsmath,amssymb,amsfonts}
\usepackage{algorithmic}
\usepackage{graphicx}
\usepackage{textcomp}
\usepackage{xcolor}
\def\BibTeX{{\rm B\kern-.05em{\sc i\kern-.025em b}\kern-.08em
    T\kern-.1667em\lower.7ex\hbox{E}\kern-.125emX}}
\newtheorem{definition}{Definition}
\usepackage{amsfonts}
\usepackage{amssymb}
\usepackage{amsmath}
\usepackage{multirow}
\usepackage{booktabs}
\usepackage{hyperref}

\begin{document}

\title{Detecting Neighborhood Gentrification at Scale via Street-level Visual Data\\
}

\author{\IEEEauthorblockN{Tianyuan Huang\IEEEauthorrefmark{1},
Timothy Dai\IEEEauthorrefmark{3}, Zhecheng Wang\IEEEauthorrefmark{1}, Hesu Yoon\IEEEauthorrefmark{2}, Hao Sheng\IEEEauthorrefmark{3},\\
Andrew Y. Ng\IEEEauthorrefmark{3}, Ram Rajagopal\IEEEauthorrefmark{1} and Jackelyn Hwang\IEEEauthorrefmark{2}}

\IEEEauthorblockA{
\IEEEauthorrefmark{1}Department of Civil and Environmental Engineering,\\
\IEEEauthorrefmark{2}Department of Sociology,\\
\IEEEauthorrefmark{3}Department of Computer Science,\\
\textit{Stanford University}\\
\{tianyuah, timdai, zhecheng, hyoon28, haosheng, ramr, jihwang\}@stanford.edu, ang@cs.stanford.edu}}

\IEEEoverridecommandlockouts
\maketitle

\begin{abstract}
Neighborhood gentrification plays a significant role in shaping the social and economic well-being of both individuals and communities at large. While some efforts have been made to detect gentrification in cities, existing approaches rely mainly on estimated measures from survey data, require substantial work of human labeling, and are limited in characterizing the neighborhood as a whole. We propose a novel approach to detecting neighborhood gentrification at a large-scale based on the physical appearance of neighborhoods by incorporating historical street-level visual data. We show the effectiveness of the proposed method by comparing results from our approach with gentrification measures from previous literature and case studies. Our approach has the potential to supplement existing indicators of gentrification and become a valid resource for urban researchers and policy makers.
\end{abstract}

\begin{IEEEkeywords}
Urban Computing, Computer Vision, Change Detection, Multi-Instance Learning
\end{IEEEkeywords}

\section{Introduction}
Gentrification is the process of reinvestment, renewal and the influx of middle- and upper-middle-class residents into previously disinvested and declined urban neighborhoods\cite{Housing2000}. Detecting gentrifying neighborhoods is crucial for investigating the dynamics of the urban sphere, with consequences in both economic and social inequality.
Traditional approaches to detecting gentrification rely on collecting demographic information, like the Decennial Census conducted by U.S. Census Bureau. However, the data produced through such survey-based methods are often restrained by their spatial and temporal granularity: the most comprehensive national census in the US takes place every 10 years, while more frequent measurements like the 1-year demographic estimates from the American Community Survey (ACS) only cover areas with populations of $65,000+$. ACS data at smaller geographies are only available as 5-year estimates due to their smaller annual sample size. 
In short, traditional methods inevitably face limitations in spatial and temporal granularity. This calls for the adoption of other sources of data to measure neighborhood gentrification in an effort to understand the dynamics of urban change.
\input{figure/title}

To overcome the aforementioned limitations and facilitate a more comprehensive measurement of the urban environment, there have been recent attempts to adopt street-level visual data in cities.
Visible aspects of gentrification express the social transformation of a neighborhood that is facilitated by a complex combination of actors \cite{hwang_gentrification}\cite{smith1986}, and street-level images offer a valuable perspective in capturing the visual characteristics of neighborhoods' built environments.
Furthermore, street-level images update at a more frequent basis compared to the slow decennial rate of census surveys; to illustrate, Google Street View has gathered historical imagery dating back to $2007$ and has updated every 1-3 years for most urban areas in the US, creating a digital time capsule of cites. \cite{hwang_gentrification} surveys Google Street View images to measure gentrification. Efforts in image segmentation and classification have been made to measure the perceptions of the quality of urban appearances and their changes with street-level visual data \cite{Naik7571}\cite{Naik2014StreetscoreP}. \cite{ilic_gentrification} detects visual property improvements from the temporal sequences of Google Street View images in Ottawa, Canada, confirming urban areas that are known to be gentrifying as well as revealing those gentrifying areas that were previously unknown. However, most previous projects are also city-specific and rely on hand-labeling large amounts of training data, hindering their models and representations' ability to transfer to multiple cities. More importantly, existing methods only detect atomic units of gentrification-related features but fail to characterize the neighborhood as a whole. Given that gentrification is defined at the neighborhood level, aggregating atomic units of features towards a neighborhood representation is crucial. As is shown in figure \ref{fig:title}, the aggregation step faces challenges regarding information dilution, because only a small percentage of all sampled street views in a neighborhood contain gentrification-related visual cues.

In this project, we propose a 2-step framework to detect gentrification at scale using street-level visual data. We conclude with an assessment of neighborhoods across three different cities experiencing widespread gentrification: Oakland, CA; Seattle, WA; and Denver, CO. Our major contribution is three-fold: 1) We introduce a framework to identify gentrification signals from large and noisy street-level visual data in a scalable manner by effectively utilizing auxiliary datasets in other modalities. 2) We develop the aggregation of atomic units of gentrification signals to the neighborhood level, where each boundary can be defined arbitrarily which enables a flexible comparison with other gentrification measures. (3) We validate our proposed approach by qualitatively examining details of potentially gentrifying neighborhoods across all three cities.

\section{Related Work}
\subsection{Neighborhood Gentrification Measures}
While most gentrification scholars rely on publicly available demographic and housing data, often from the Census and ACS, a handful of studies have systematically examined the visual features of gentrification. To measure gentrification in cities at a large scale, field surveys were adopted on the city block level where raters walked through neighborhoods to record visible cues such as renovation and reinvestment of buildings\cite{census1996}. More recently, human coders examined Google Street View images to detect theoretically driven indicators of gentrification\cite{hwang_gentrification}. \cite{ilic_gentrification} used deep neural networks and siamese networks with street view images to map property improvement in the city of Ottawa, and the author collected crowd-sourced labels of pairwise images via website questionnaires similar to \cite{naik2014streetscore}. Apart from that, business activities data were also used to quantify urban change and measure gentrification\cite{Meltzer2016GentrificationAS}\cite{Glaeser2018NowcastingGU}. Fueled by more densely-sampled Google Street View images in multiple cities, together with the neighborhood representation learned via the attention mechanism, our proposed framework provides a more convenient and comprehensive tool to detect gentrification at scale.

\subsection{Street-level Visual Data}
Street-level visual data have been utilized in a variety of application scenarios in cities, such as indicating regional functions \cite{Gong2019ClassifyingSS}, examining environmental associations with chronic health outcomes \cite{Nguyen2021GoogleSV} and measuring populace’s well-being \cite{Lee2021PredictingLI}. They have also been adopted in spatial-temporal representation learning \cite{wang2020urban2vec}\cite{jenkins2019unsupervised}\cite{tian2021} to produce region embedding for urban neighborhoods. Recent research have shown the possibility to infer socioeconomic attributes such as income, race, education, and voting patterns from street view data \cite{gebru2017using}. Furthermore, \cite{naik2014streetscore} aimed to predict the perceived level of safety on the streets from their visual data, and \cite{Naik7571} developed a similar mechanism to analyze pairwise images taken in $2007$ and $2014$ respectively to map out physical improvement and decline, and correlate with social-economic attributes. However, existing methods aggregate the coordinate-level signals to neighborhood-level representations through the mean operator, which can potentially dilute sparsely-distributed key elements.

\subsection{Multiple-Instance Learning}
Multiple-instance learning is a type of supervised learning where a single class label is assigned to a bag of instances. In the simple case, a bag is labeled positive if there is at least one instance in it which is positive. Traditionally, Multiple-instance learning approaches rely on the mean pooling or the max pooling when aggregating instances to the bag \cite{Feng2017DeepMN}\cite{Pinheiro2015FromIT}. Fully-connected neural networks were adopted and proved to be beneficial\cite{Wang2018RevisitingMI}, and the recent work on incorporating a gated attention mechanism \cite{Ilse2018AttentionbasedDM} showed that such a mechanism outperforms commonly used pooling operators in multiple-instance learning. We follow this line of research since it allows a learnable way to assign weights to instances. However, our problem setup has a more complex case where a bag may be still labeled as negative if all the instances in it are not negative (e.g. A non-gentrifying neighborhood with very few signs of improvement). Our proposed method utilizes both the bag labels and instance labels and bridges them with the gated attention mechanism.

\input{figure/method}
\section{Methods}
\subsection{Problem Statement}
\begin{definition}
[Time-lapsed Street View Pairs] Each time-lapsed street view pair consists of $2$ images capturing the same street-level scene $t^{(i)}=(s^{(i)}_{e}, s^{(i)}_{l})$, where $s^{(i)}_{e}$ is recorded at an earlier timestamp than $s^{(i)}_{l}$.
\end{definition}
Time-lapsed street-level images in urban neighborhoods capture the change in physical appearance of the built environment, which can uncover relevant signals of gentrification.
\begin{definition}
[Neighborhood Container] A metropolitan area consists of a set of urban neighborhoods $N=\{n_1, n_2, \dots, n_N\}$ with disjoint geographical geometries. Each neighborhood unit $n_j$ contains a set of time-lapsed street view pairs $n_j=\{t^{(1)}_{j},t^{(2)}_{j},\cdots,t^{(K)}_{j}\}$.
\end{definition}
Our goal is to identify gentrifying neighborhoods in cities in an accurate, scalable and explainable approach. To do so, we propose a 2-step method as is shown in Figure \ref{fig:method}, each step having its own training data and labels, as follows:
\begin{itemize}
    \item Step 1: Detect new constructions and major renovations of buildings from time-lapsed street view pairs within each neighborhood.
    \item Step 2: Aggregate street view representations to the neighborhood level and predict the neighborhood's gentrification status.
\end{itemize}
\subsection{Step 1. Change Detection on Street Views}
Because new constructions and renovations of buildings are widely considered characteristics of gentrifying neighborhoods, it follows that our first step is to formulate a change detection task for time-lapsed street view pairs. To define such a task, we label each time-lapsed street view pair $t_i$ with a binary class: positive for pairs that contain meaningful change (e.g. new construction) and negative for pairs that do not. Here, ``meaningful change'' signals gentrification, while an absence of meaningful change includes either simply no change or random changes like cars on the road or weather conditions which add potentially noisy signals.

For this task, we develop a Siamese network \cite{Koch2015SiameseNN} with $L+1$ layers, which includes a twin ResNet\cite{resnet} module to realize a non-linear embedding from the input domain--- time-lapsed street view pairs--- to some Euclidean spaces $\mathbb{R}^d$. 
Let $\mathbf{h}^{(i)}_{e, L}$ represent the $(L)$th hidden vector for the street view with an earlier timestamp $s^{(i)}_e$ in the time-lapsed street view pair $t^{(i)}$ and $\mathbf{h}^{(i)}_{l, L}$ denote the same for the later street view $s^{(i)}_l$. We notice that a naive element-wise distance metric of the hidden vectors alone is insufficient in learning street-level change from time-lapsed street view pairs. Our concern follows the fact that most street-level changes correlate to the context of the scene at a certain level (e.g. the camera angle). Therefore, we propose the final $(L)$th hidden vector for the time-lapsed street view pair $t^{(i)}$ as the concatenation of both images' hidden vectors and their distance, represented by their element-wise difference, yielding ${h}^{(i)}_{L}\in\mathbb{R}^M$ as follows:
\begin{equation} \label{eq: sia_concat}
\mathbf{h}^{(i)}_{L} = \left[(\mathbf{h}^{(i)}_{l, L}-\mathbf{h}^{(i)}_{e, L})^{\top}, (\mathbf{h}^{(i)}_{l, L})^\top, (\mathbf{h}^{(i)}_{e, L})^\top\right]^{\top},
\end{equation} where $M=3d$.
The final vector representation is followed by a fully-connected layer, which is given to a single output unit with the sigmoid activation. More precisely, the prediction scalar is given as:
\begin{equation}
\mathbf{p}(t^{(i)}) = \sigma({\alpha}^\top(\mathbf{h}^{(i)}_{L})).
\end{equation}
Let $\mathbf{y}(t^{(i)})$ be the label for the time-lapsed street view pair $t^{(i)}$. We assume $\mathbf{y}(t^{(i)})=1$ when there is meaningful change in $t^{(i)}=(s^{(i)}_{e}, s^{(i)}_{l})$ and $\mathbf{y}(t^{(i)})=0$ otherwise. We adopt binary cross-entropy loss in our objective function in the following form:
\begin{multline} \label{eq: bce}
    \mathcal{L_S}(t^{(i)}) = \mathbf{y}(t^{(i)})\log\mathbf{p}(t^{(i)})+\\ (1-\mathbf{y}(t^{(i)}))\log(1-\mathbf{p}(t^{(i)})).
\end{multline}

\subsection{Step 2. Neighborhood-level Aggregation}
A neighborhood $j$ has a set of $K$ time-lapsed street view pairs $n_j$, and there is a single binary label $Y$ associated with $n_j$ indicating whether it is gentrifying or not. We acquire the physical boundaries of what we call a ``neighborhood'' from government-defined census tracts (using 2010 boundaries), as well as its binary label $Y$ derived from publicly available ACS data \cite{Hwang2020GentrificationWS}. We assume neither ordering nor dependency within the bag $n_j$, so we propose to aggregate street view embedding in $n_j$ through multi-instance pooling. Specifically, we extract the embedding of each time-lapsed street view pair $\mathbf{h}^{(i)}_{L}$ for each $t_j^{(i)}$ in $n_j$ from the change detection model, then calculate neighborhood-level feature vectors $\mathbf{n}_j$ by taking the weighted average of the instance embeddings for each time-lapsed street view pair:
\begin{equation} \label{eq: agg}
    \mathbf{n}_j = \sum_{i=1}^{K}a_{i}\mathbf{h}^{(i)}_{L}.
\end{equation}
\\
\textbf{Gated attention mechanism} We notice that a portion of instances in each bag are more significant than others when predicting the bag's label. For example, a time-lapsed street view pair with radical residential renovations delivers strong signals of gentrification, while a pair of nearly identical street views capturing a nondescript highway scene cannot provide as much knowledge about a neighborhood's gentrification status as the former. If we use the mean operator when aggregating (i.e., $a_i=\frac 1 K$), the signal from significant instances might become diluted in the neighborhood-level representation.
Therefore, we propose the adoption of a gated attention mechanism \cite{multi} by parameterizing $a_i$ in equation \ref{eq: agg} as follows:
\begin{equation}
    a_i = \frac{\exp\{\mathbf{w}^\top(\mathrm{tanh}(\mathbf{V}\mathbf{h}^{(i)}_{L})\odot\mathrm{sigm}(\mathbf{U}\mathbf{h}^{(i)}_{L}))\}} {\sum_{j=1}^{K}\exp\{\mathbf{w}^\top(\mathrm{tanh}(\mathbf{V}\mathbf{h}^{(j)}_{L})\odot\mathrm{sigm}(\mathbf{U}\mathbf{h}^{(j)}_{L}))\}},
\end{equation}
where $\mathbf{w}\in \mathbb{R}^{W\times1}$, and $\mathbf{U}, \mathbf{V}\in \mathbb{R}^{W\times M}$ are learnable parameters. Notably, the gated attention layer adds non-linearity when learning the weight $a_i$ for each instance $\mathbf{h}^{(i)}_{L}$. Our proposed model aims to assign different weights to each time-lapsed street view pair so that the most significant street view-level representations inform the neighborhood's gentrification status.

Finally, the neighborhood representation is followed by a classifier with a fully-connected layer and sigmoid activation as follows:
\begin{equation}
\mathbf{P}(n_{j}) = \sigma({\beta}^\top(\mathbf{n}_{j})).
\end{equation}
Again, we use a cross-entropy loss function to train such a classifier, where $\mathbf{Y}(n_j)=1$ when $n_j$ is a gentrifying neighborhood, and $\mathbf{Y}(n_j)=0$ when $n_j$ is non-gentrifying:
\begin{multline} \label{eq: bce_n}
    \mathcal{L_N}(n_j) = \mathbf{Y}(n_j)\log\mathbf{P}(n_j)+\\ (1-\mathbf{Y}(n_j))\log(1-\mathbf{P}(n_j)).
\end{multline}

\section{Experiments}
To demonstrate the effectiveness of our framework, we conduct experiments for the cities of Oakland, California; Seattle, Washington; and Denver, Colorado. We adopt the census tracts defined by the US Census Bureau as the unit of neighborhoods, since the referred gentrification labels \cite{Hwang2020GentrificationWS} are readily available under these same neighborhood delineations. Despite these predefined units, our framework is flexible when applying to other geographic units of neighborhoods (e.g. census block groups) or even customized boundaries.

\subsection{Datasets}
\textbf{Historical Google street views} The time-stamped street view images are obtained from Google Static Street view API\footnote{Available at \url{https://developers.google.com/maps/documentation/streetview}} between the years of $2007$ and $2022$ for the aforementioned cities: Oakland, Seattle and Denver. To select the geospatial locations for downloading those images, we sample one coordinate every $50$ to $100$ meters along the road network, Table \ref{table:sv_stats} shows the number of street view images we sampled for each city. For each time-lapsed street view pair, we ensure that the earlier image is captured no later than $2010$ and the later image is captured no earlier than $2018$ in order to maximize the occurrences of observable urban change within the time interval. Figure \ref{fig:sv_map} shows the geospatial distribution of those street views.
\input{figure/sv_map}

\textbf{Construction permits} Construction permit data are fetched from the online permit center of the corresponding city government. Each row of permit data includes information of the permit's issued date, category, geospatial coordinate and cost estimate, among other miscellaneous government-mandated information. As a data preprocessing step, we adjust the dollar job values for inflation, keep only the ``new'', ``alteration'' and ``addition'' categories and remove construction jobs with a total cost of less than $\$60,000$ in a single year. This way, urban change recorded by our filtered permits is significant enough to be viewed as potential signals of neighborhood gentrification.

\input{figure/permit}
\textbf{Business directories} Historical business data are licensed from Data Axle's ReferenceUSA\footnote{Available at \url{http://www.referenceusa.com/}} product, in which businesses are recorded each year by their name, NAICS classification, address, etc. Here, we extract businesses that have converted from an essential retail business (e.g., laundry, grocery stores) into a discretionary retail business (e.g., high-end restaurants, coffee shops, art galleries, etc.), specifically between the years of $2007$ and $2020$. We consider such business changes as another source of gentrification signals.

\textbf{Gentrification measures} For comparison, we use gentrification measures derived from social and demographic data obtained via the 2005-2009 and 2015-2019 ACS data, following the approach used in \cite{Hwang2020GentrificationWS}. The measures include $3$ categories for census tracts: gentrifying, non-gentrifying and non-gentrifiable. Non-gentrifiable neighborhoods are affluent neighborhoods at the beginning of the period and are thus ineligible to gentrify. As a point of clarification, gentrifiable neighborhoods include both gentrifying and non-gentrifying neighborhoods, and only gentrifying neighborhoods experience gentrification according to the measures. Our work focuses on gentrifiable neighborhoods in cities; we aim to detect all gentrifying neighborhoods from the gentrifiable neighborhoods. Table \ref{table:sv_stats} shows the number of gentrifying and non-gentrifying neighborhoods (i.e., census tract) in each of our studied cities.
\input{table/sv_stats}

\subsection{Training Details}
\textbf{Change detection model} Our experiments utilize ResNet18 \cite{resnet} as the backbone network to transform our input from the image space to the vector space of $\mathbb{R}^d$ where $d=512$, thereby setting $M=512\times3$ as the dimension of $\textbf{h}_{L}^{(i)}$. Furthermore, the threshold of $0.5$ is adopted in the binary predictor $\mathbf{p}(t^{(i)})$. According to the geospatial coordinates from construction permits and business directories data, $2,476$ street view images (i.e., $1,238$ time-lapsed street view pairs) are sampled to train the change detection model. For each valid coordinate, we download $2$ time-lapsed street view pairs: one pair labeled as positive (i.e., $\mathbf{y(t^{(i)})}=1$) which comprises $2$ images taken before and after the date of the change respectively, and the other as negative (i.e., $\mathbf{y(t^{(i)})}=0$) where both images are captured before the date of change. Finally, we split the dataset into a training set ($70\%$) and a test set ($30\%$). 

\textbf{Gated attention model} We sample $K=100$ to $200$ pairs of time-lapsed street views for each census tract depending on its size and road density. $W =128$ is used when setting the shape of the matrices $\mathbf{V}$, $\mathbf{U}$ and $\mathbf{w}$. Similar to the change detection model, we split the neighborhood dataset into a training set ($70\%$) and a test set ($30\%$), and we adopt the threshold of $0.5$ in the binary predictor $\mathbf{p}(t^{(i)})$. The labels in this step indicate whether the neighborhood $n_{j}$ is gentrifying (i.e., $\mathbf{Y}(n_{j})=1$) or not (i.e., $\mathbf{Y}(n_{j})=0$).
\input{figure/weight_full}
\subsection{Baselines}
We evaluate our method with three ablation studies. 1) \textbf{Pre-trained \& no attention} skips both the change detection model in Step 1 and the attention mechanism in Step 2. Instead, it generates image embeddings via the backbone model pre-trained on ImageNet and uses the mean operator when aggregating instance embedding in Step 2 (i.e., $\mathbf{n}_j = \sum_{i=1}^{K}\frac{1}{K}\mathbf{h}^{(i)}_{L}$). 2) \textbf{No attention} keeps the same training setting in Step 1, skips the attention mechanism in Step 2 and uses the mean operator in the aggregation step similar to \cite{Naik7571}. 3) \textbf{E2E} drops the change detection labels in Step 1 and is trained in an end-to-end manner from pairwise image input to neighborhood prediction output with only gentrification attributes as labels. These three models serve as our baselines.

\section{Results and Discussion}
\subsection{Predicting Gentrification Attributes}
We first train and evaluate the change detection model as described in Step 1. The siamese-based twin network achieves a $93\%$ accuracy in predicting the positive pairs obtained through permits and business directories. Next, we extract the time-lapsed street view pairs' embeddings and plug them into Step 2 of the full model as well as \textbf{No attention} to benchmark how well each model predicts the neighborhood gentrification attributes from the measurement in \cite{hwang_gentrification}. To offset the class imbalance of gentrifying and non-gentrifying labels, we report the balanced accuracy of the model performance on the test set. As shown in Table \ref{table:res}, our full model outperforms all the baselines by a significant margin. Specifically, \textbf{Pre-trained \& no attention} suffers from loss oscillation during training, and it predicts test neighborhoods to be either all non-gentrifying or all gentrifying in the three studied cities. \textbf{No attention} misclassifies a portion of non-gentrifying neighborhoods in Oakland's test set, and it predicts all neighborhoods to be gentrifying in Seattle and Denver. \textbf{E2E} experiences overfitting and generalizes poorly on the test set in all three cities, possibly due to the fact that a certain amount of noisy signals in street views are detected as gentrification-related cues since \textbf{E2E} drops the change labels in Step 1.
In these ablation studies, we demonstrate that both Step 1 and Step 2 are vital in extracting enough urban change information from the time-lapsed street view pairs to predict gentrification status. Our full model achieves $74\%$ balanced accuracy and $89\%$ recall across the three cities, predicting more neighborhoods to be gentrifying compared to the measurement labels. However, we note that the gentrification measurement \cite{hwang_gentrification} relies on demographic and housing data from the ACS, while our approach leverages a different data source: the physical appearance of cities captured by street-level imagery. Thus, to further validate our proposed model, we perform qualitative analysis and case studies.
\input{table/res}
\subsection{Interpreting the Learned Weights}
To uncover how our proposed method distinguishes gentrifying from non-gentrifying neighborhoods, we examine the attention mechanism closely by visualizing the weight distribution for all neighborhood containers. In Figure \ref{fig:weight_full}, we sample $K=100$ time-lapsed street view pairs for each neighborhood $t^{(i)}$ and sort the pairs based on their value of $a_i$ in descending order. We observe that the curve for non-gentrifying neighborhoods drops off much faster than the curve for gentrifying neighborhoods. This means that most time-lapsed pairs in non-gentrifying neighborhoods are assigned an $a_i$ value close to an average value of $0.01$ with only a small number of highly weighted pairs. In contrast, the gentrifying neighborhoods have many more highly weighted pairs and thus can be characterized by a more polarized distribution of $a_i$. We intuit this observation with the logical assumption that gentrifying neighborhoods will contain more time-lapsed street view pairs with significant changes. With this difference in the learned weights between gentrifying and non-gentrifying neighborhoods, we show that our proposed model can extract potential signals of gentrification over random noise by assigning significant weights in order to classify gentrifying neighborhoods. 

To further demonstrate the semantic meaning of the learned weights, we sample $20$ time-lapsed street view pairs corresponding to the highest and lowest weights in a randomly sampled neighborhood and visualize their street views in Figure \ref{fig:weight_full}. We find that pairs depicting buildings have higher weights, whereas pairs of highways and vegetation have lower weights. This validates that the attention model prioritizes high signaling pairs and deprioritizes pairs that might add noise or provide uninformative signals, ultimately avoiding information dilution by filtering out less relevant instances via lower weights.

\input{figure/case_map}
\input{figure/cases}
\subsection{Case Studies: Detecting Potential Gentrifying Neighborhoods}
Given that our proposed measurement of gentrification leverages a different data source and predicts more neighborhoods to be gentrifying compared to the measurement labels, we seek to explore ways to supplement the existing labels in order to develop a more comprehensive and flexible measurement of gentrification. In particular, we explore the following questions: Which neighborhoods are being classified as gentrifying yet labeled as non-gentrifying? Can we find signals of gentrification in those neighborhoods? Could they be in the early stages of gentrification? To explore these questions, we visualize the prediction results on all gentrifiable neighborhoods and overlay the gentrification labels derived from \cite{hwang_gentrification} as shown in Figure \ref{fig:case}. At first glance, we observe that many neighborhoods in the discrepancy class are adjacent to neighborhoods labeled as gentrifying, leading us to surmise that such neighborhoods could possibly be in the early stages of gentrification. The idea is that the socioeconomic effects of gentrification in already-gentrifying neighborhoods may spillover into surrounding neighborhoods\cite{Guerrieri2010EndogenousGA}.

To further evaluate the prediction discrepancy, we select four census tracts across the 3 cities to examine image-by-image as further case studies. These tracts are located in the neighborhoods of Prescott, Oakland; McClymonds, Oakland; Beacon Hill, Seattle; and Lincoln Park, Denver. Figure \ref{fig:cases} shows that our proposed method identifies a series of gentrification signals by assigning higher weights for those signaling street view pairs. Specifically, we observe some major changes in residential buildings including repairing and repainting of houses as well as new construction of single family houses and apartment buildings. Moreover, the rehabilitation of vegetation and greenery and refurbishment of infrastructure such as roads and sidewalks are also detected. Since the gentrification measurement developed through ACS data uses 5-year estimates of social-economic metrics across census tracts, local changes in the built environment can easily be overlooked. Hence, our proposed method can supplement the existing measurement by identifying specific signals of physical neighborhood enhancement which are crucial to the gentrification process and identifying specific locations within census tracts where gentrification is occurring.

\section{Conclusion and Discussion}
In this work, we propose a framework to detect gentrifying neighborhoods at scale by applying computer vision and statistical analysis to street-level visual data. We have --- for the first time --- detected and aggregated the atomic units of gentrification signals to the neighborhood level through a learnable mechanism and validated this mechanism by predicting gentrification attributes in multiple cities with a substantially larger number of neighborhoods compared to previous attempts. By examining on-the-ground examples of gentrification-related visual cues, we observe potentially gentrifying neighborhoods and evaluate them through several case studies. Given the fact that gentrification scholars still lack a consensus on the best way to measure gentrification in a quantitative way \cite{BrownSaracino2017ExplicatingDA}\cite{Zuk2015GentrificationDA}\cite{SchnakeMahl2020GentrificationNC}, our proposed approach has demonstrated its potential to serve as a valid resource to supplement and refine existing approaches. In the three cities we analyze, we find that neighborhoods which are adjacent to already-gentrifying neighborhoods show evidence of gentrification in terms of visual cues, indicating that such neighborhoods could potentially be in the early stages of gentrification or that gentrification is only occurring in subsections of the neighborhood.

While our data-driven method provides a novel approach to the measurement of neighborhood gentrification, it is still subjective to the following limitations: 1) Our labeling of visual cues of gentrification relies on auxiliary datasets (i.e., permits and business data) which are smaller compared to the image dataset we finally deploy, thus facing challenges in terms of out-of-distribution data on the test set. Specifically, random noisy signals in time-lapsed street view pairs (e.g., perspective discrepancies, camera angle blocks) may affect our model's ability to generalize in different scenarios correctly. 2) Since street-level image data like building construction and renovation are among the most apparent visual signals of gentrification, our model may assign more attention to the places with a higher presence of buildings while neglecting the less populated ones. 3) Finally, demographic and economic attributes are still necessary to take into account when measuring gentrification in a comprehensive way in order to avoid possible visual biases. Despite these limitations, we believe our proposed approach and analysis provide a flexible framework that can help guide scholars and policy makers in developing quantitative measures of gentrification. More broadly, locating and predicting gentrification-like urban change can benefit local governments and planning agencies via identifying places at risk of displacement and helping them prioritize certain infrastructure investments and target policy interventions. In future work, our pipeline can be extended to more downstream tasks of detecting relevant elements and signals in other aspects of urban change.

\section{Acknowledgement}
This project was supported by the Google Cloud Grant from the Stanford Institute for Human-Centered Artificial Intelligence. The author would like to thank Sarthak Kanodia, Herman Donner and Jeremy Irvin for their extensive guidance.

\bibliographystyle{IEEEtran}
\bibliography{ref}

\end{document}

%% file: figure/title.tex
\begin{figure}[ht!]
  \centering
  \includegraphics[width=0.48\textwidth]{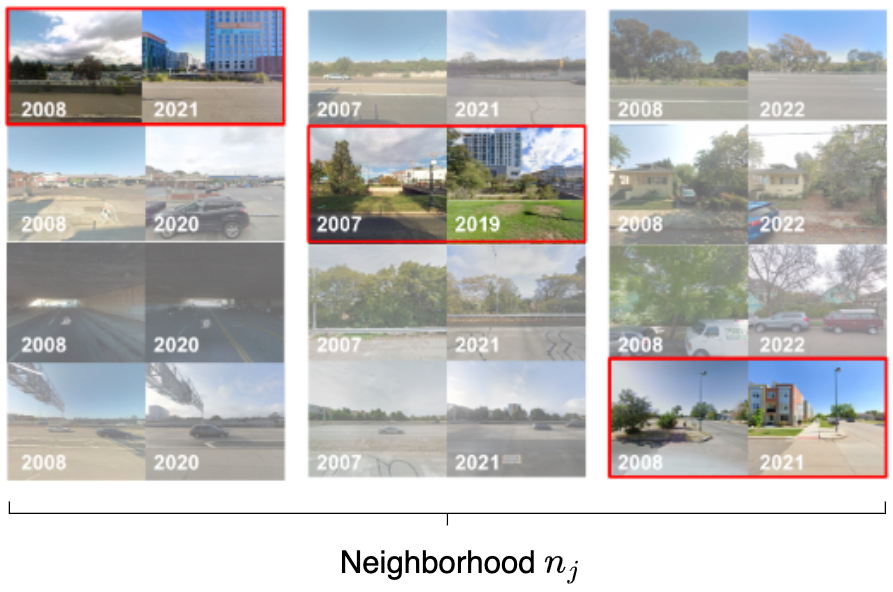} 
  \caption{A neighborhood container with time-lapsed street view pairs. For each neighborhood, our proposed framework detects gentrification visual cues (highlighted pairs) from the sampled street-level visual data and then learns the aggregated neighborhood representation to classify its gentrification status.}
  \label{fig:title}
\end{figure}

%% file: figure/method.tex
\begin{figure*}[ht!]
  \centering
  \includegraphics[width=0.93\textwidth]{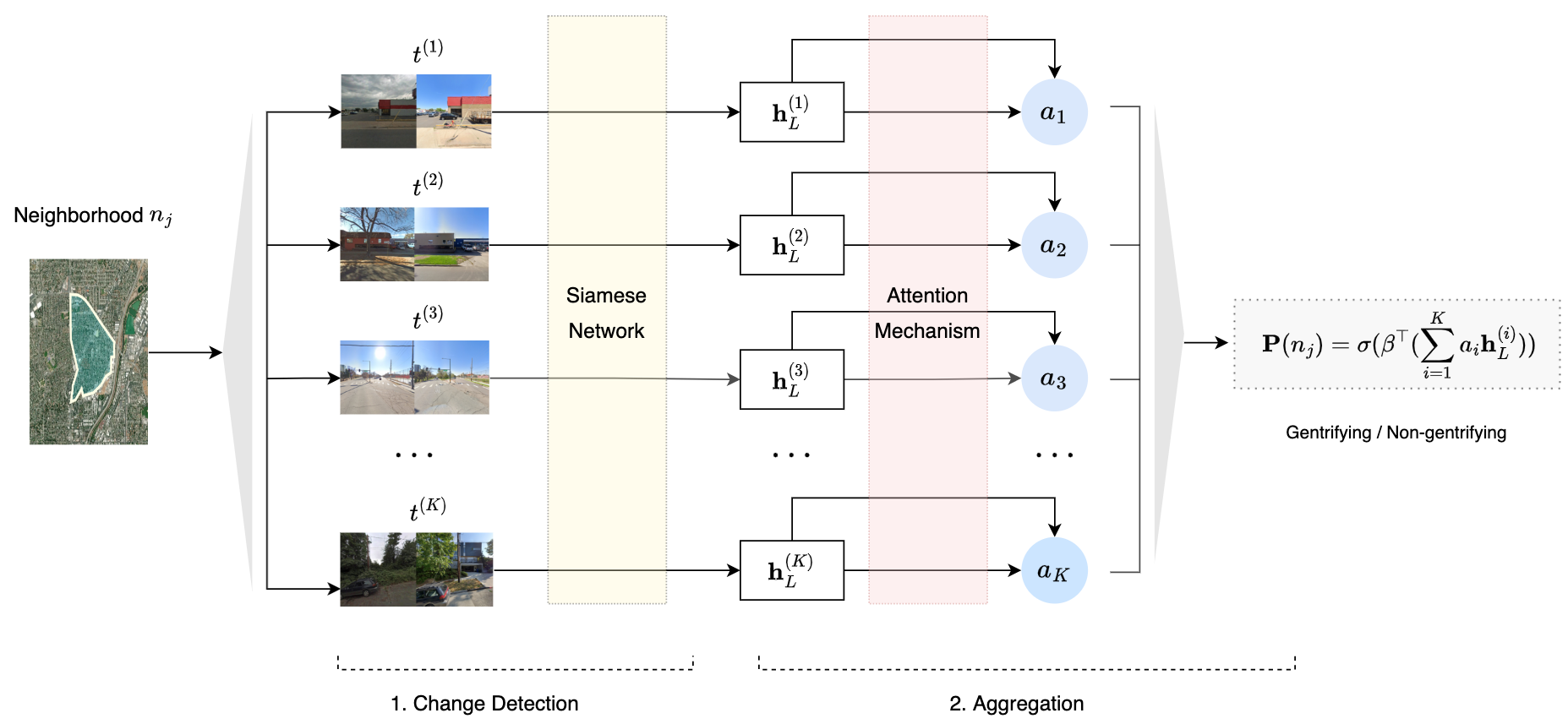} 
  \caption{Overview of the proposed method. Step 1 detects new constructions and renovations of buildings from time-lapsed street view pairs via Siamese-based networks. Step 2 aggregates street view representations to the neighborhood level with the attention mechanism to predict the gentrification status.}
  \label{fig:method}
\end{figure*}

%% file: figure/sv_map.tex
\begin{figure}[ht!]
  \centering
  \includegraphics[width=0.48\textwidth]{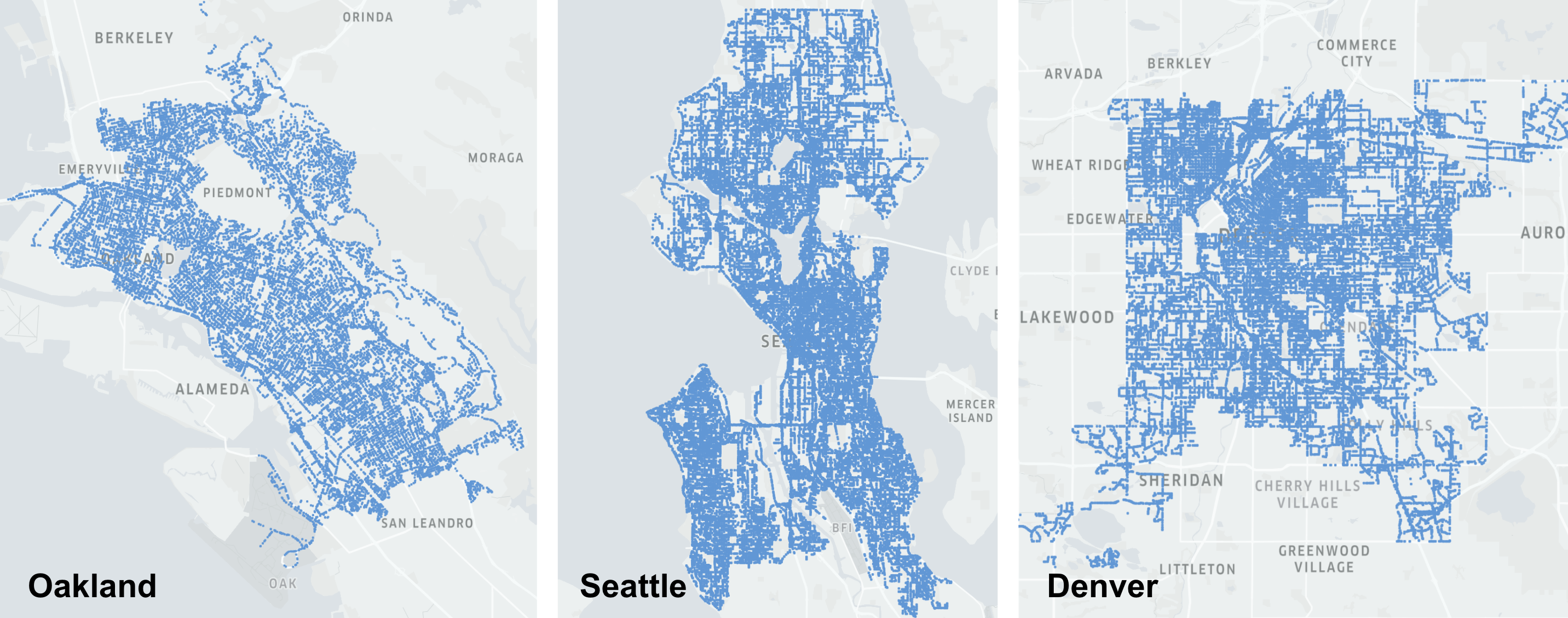} 
  \caption{Geospatial distribution of sampled street views in three studied cities.}
  \label{fig:sv_map}
\end{figure}

%% file: figure/permit.tex
\begin{figure}[ht!]
  \centering
  \includegraphics[width=0.48\textwidth]{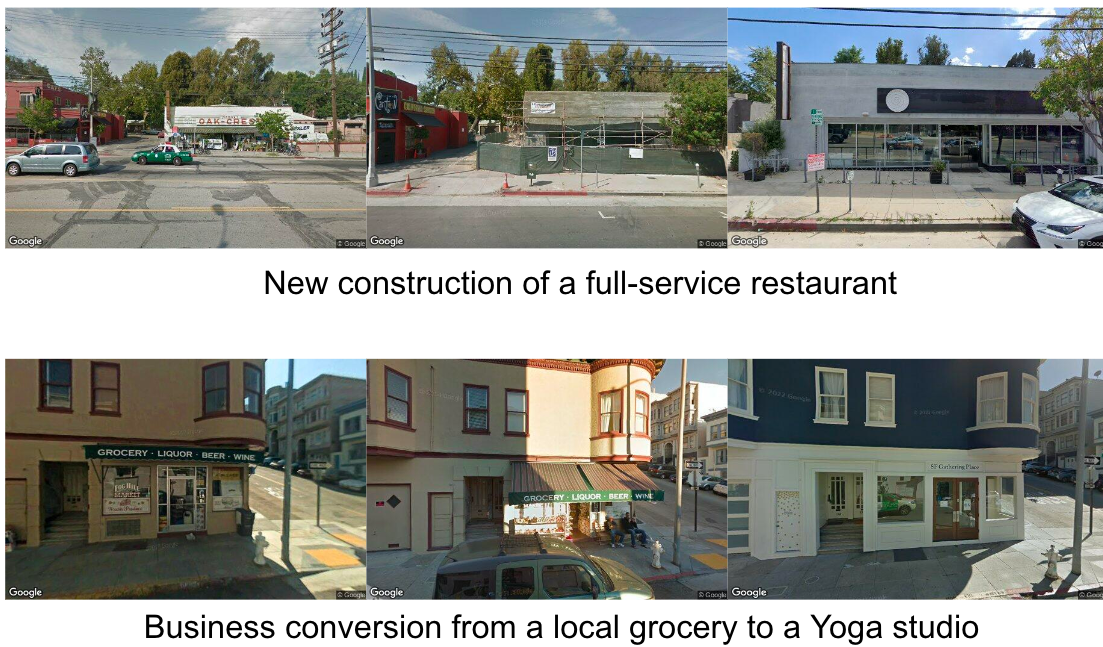} 
  \caption{Examples of gentrification-related visual cues obtained through construction permits and business directories.}
  \label{fig:permit}
\end{figure}

%% file: table/sv_stats.tex
\begin{table}[ht!]
    \caption{Dataset statistics}
    \centering
    \resizebox{0.45\textwidth}{!}{%
        \begin{tabular}{l|c|c|c}
          \toprule
            \multirow{2}{*}{City} & \multirow{2}{*}{\# Street views} & \multicolumn{2}{|c}{\# Neighborhoods (census tracts)} \\ 
            \cmidrule{3-4}&
            &  Gentrifying  &  Non-gentrifying  \\
          \midrule
          Seattle &$81,676$ & $37$  & $26$ \\ 
          Oakland &$41,796$ & $23$ & $31$\\ 
          Denver &$72,620$ & $47$ & $22$\\
          \bottomrule
        \end{tabular}
    }
    \label{table:sv_stats}
\end{table}

%% file: figure/weight_full.tex
\begin{figure*}[ht!]
  \centering
  \includegraphics[width=1\textwidth]{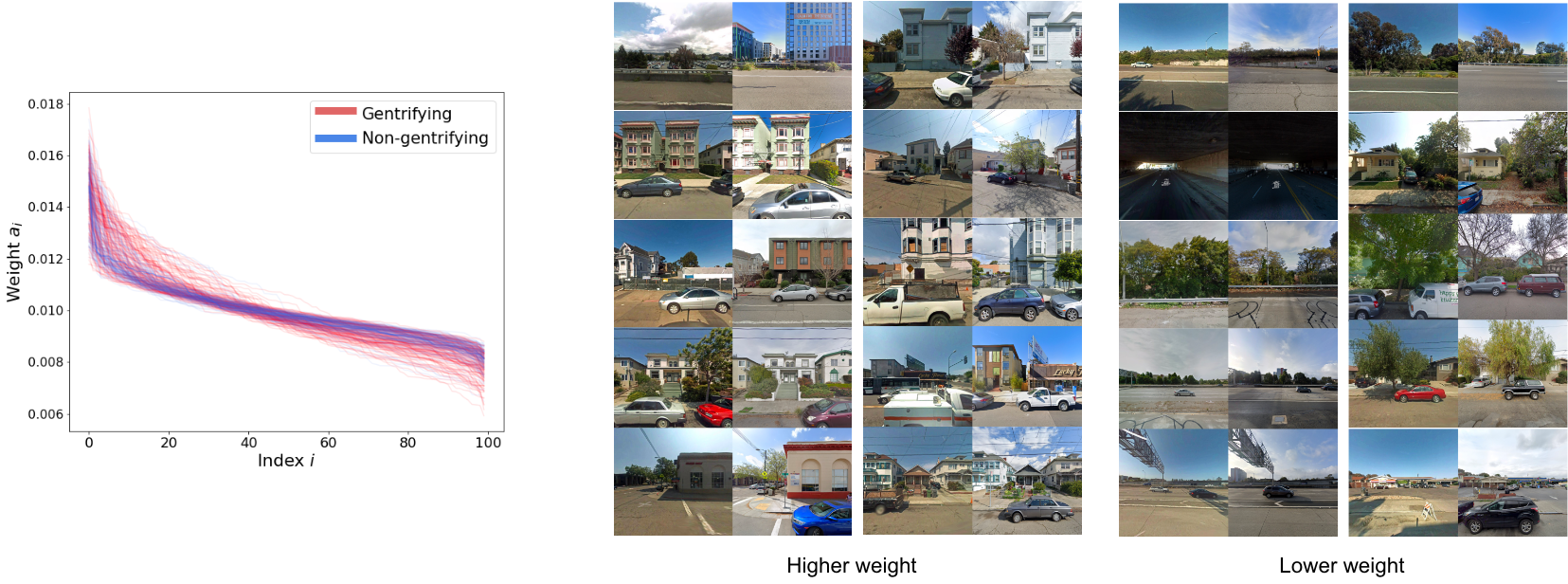} 
  \caption{{\bf Left}: weight distribution comparison between gentrifying (shown in red) and non-gentrifying (shown in blue) neighborhoods, where each line represents a neighborhood with $100$ time-lapsed street view pairs. The horizontal axis is the index of each time-lapsed street view pair $t^{(i)}$ sorted by the value of weight $a_i$ in descending order. The vertical axis is the value of weight $a_i$. {\bf Right}: Time-lapsed street view pairs with higher weights and lower weights.}
  \label{fig:weight_full}
\end{figure*}

%% file: table/res.tex
\begin{table}[ht!]
    \caption{Results of neighborhood gentrification attributes prediction}
    \centering
    \resizebox{0.48\textwidth}{!}{%
        \begin{tabular}{l|c|c|c}
          \toprule
            \textbf{Eval City} &  \textbf{Model}  &  \textbf{Acc.} & \textbf{Balanced Acc.} \\
          \midrule
          Oakland & Pre-trained \& no attention &$0.56$ & $0.50$ \\ 
          Oakland & No attention &$0.63$ & $0.57$ \\
          Oakland & E2E &$0.56$ & $0.50$ \\
          Oakland & Full model &$\mathbf{0.81}$ & $\mathbf{0.83}$ \\
          \midrule
          Seattle & Pre-trained \& no attention &$0.58$ & $0.50$ \\ 
          Seattle & No attention &$0.58$ & $0.50$ \\
          Seattle & E2E &$0.41$ & $0.35$ \\
          Seattle & Full model &$\mathbf{0.71}$ & $\mathbf{0.66}$ \\
          \midrule
          Denver & Pre-trained \& no attention &$0.68$ & $0.50$ \\ 
          Denver & No attention &$0.68$ & $0.50$ \\
          Denver & E2E &$0.63$ & $0.51$ \\
          Denver & Full model &$\mathbf{0.74}$ & $\mathbf{0.72}$ \\
          \midrule
          \midrule
          Avg of all & Pre-trained \& no attention &$0.61$ & $0.50$ \\ 
          Avg of all & No attention &$0.63$ & $0.52$ \\
          Avg of all & E2E &$0.54$ & $0.45$ \\
          Avg of all & Full model &$\mathbf{0.75}$ & $\mathbf{0.74}$ \\
          \bottomrule
        \end{tabular}
    }
    \label{table:res}
\end{table}


%% file: figure/case_map.tex
\begin{figure*}[ht!]
  \centering
  \includegraphics[width=0.8\textwidth]{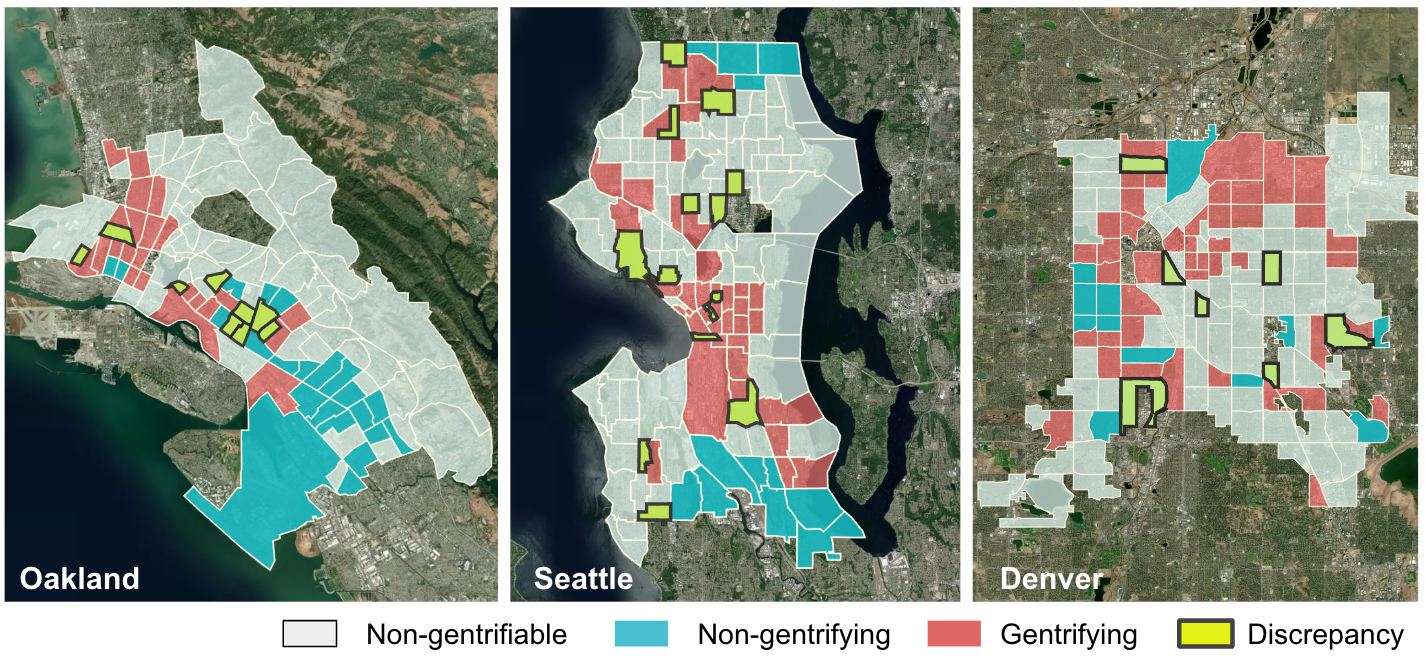} 
  \caption{Neighborhood gentrification labels and prediction discrepancy in the studied cities. Discrepancy highlights those "false positive" neighborhoods which are labeled as non-gentrifying while predicted to be gentrifying from our proposed model.}
  \label{fig:case}
\end{figure*}

%% file: figure/cases.tex
\begin{figure*}[ht!]
  \centering
  \includegraphics[width=0.95\textwidth]{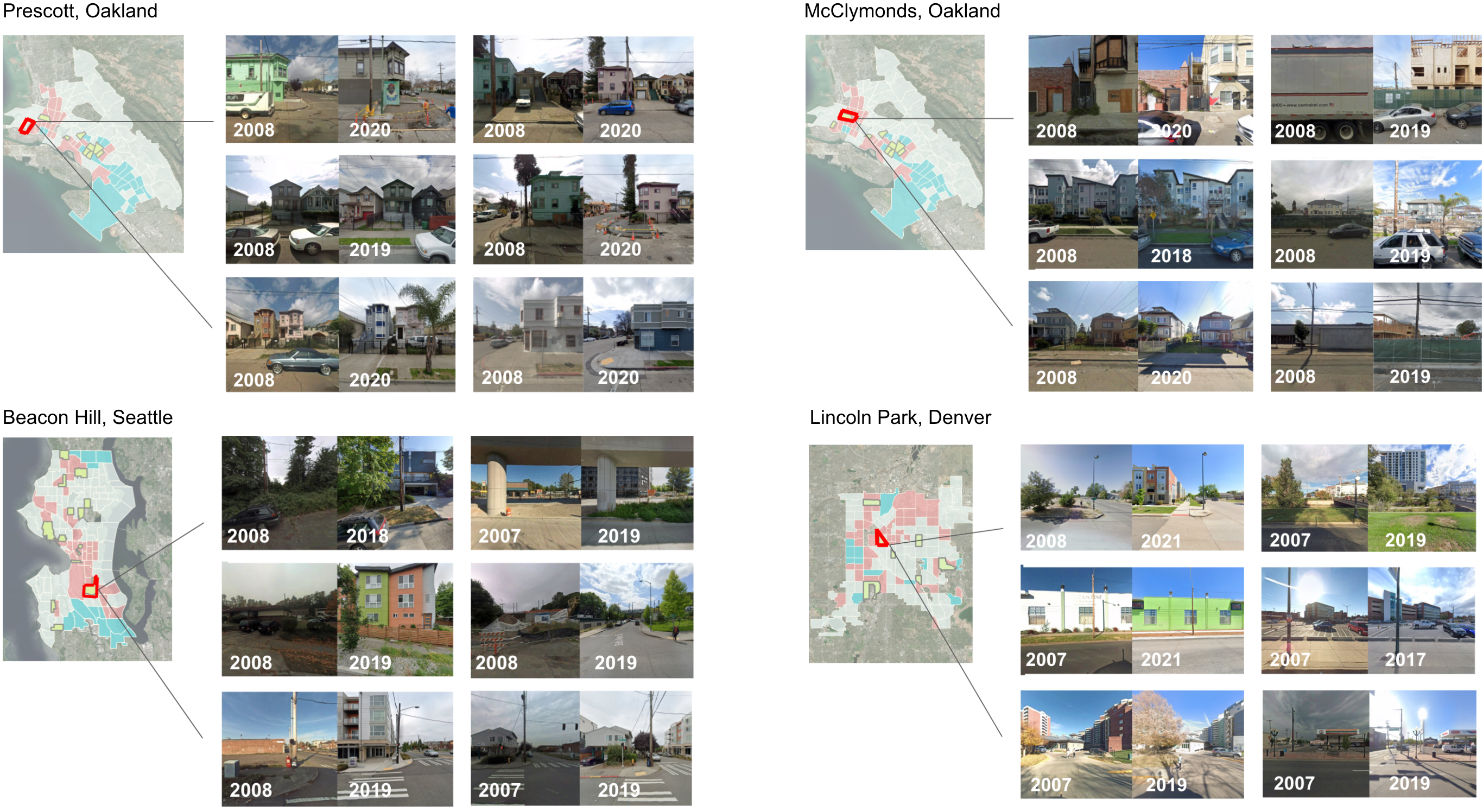} 
  \caption{Discrepancy case studies. Through assigning higher weights to the selected time-lapsed street view pairs, our proposed model detects potential gentrification signals in those neighborhoods labeled as non-gentrifying.}
  \label{fig:cases}
\end{figure*}